\documentclass[runningheads]{llncs}
\usepackage{graphicx}
\usepackage{epstopdf}
\usepackage{epsfig}
\usepackage{algpseudocode}
\usepackage{multicol}
\usepackage{multirow}
\usepackage{booktabs}
\usepackage{threeparttable}
\usepackage{enumerate}       
\usepackage{nicefrac}       
\usepackage{microtype}      
\usepackage{epstopdf} 
\usepackage{amsfonts}

\begin{document}
\title{Recurrent Deconvolutional Generative Adversarial Networks with Application to Text Guided Video Generation}

%
%
\author{Hongyuan Yu\inst{1,2}\orcidID{0000-0003-4208-1200} \and
Yan Huang\inst{1,2}\orcidID{0000-0002-8239-7229} \and
Lihong Pi\inst{3}\orcidID{0000-0003-1816-6220} \and Liang Wang\inst{1,2,4}\orcidID{0000-0001-5224-8647}}
\authorrunning{F. Author et al.}
%
\institute{University of Chinese Academy of Sciences (UCAS) \and
Center for Research on Intelligent Perception and Computing (CRIPAC), \\
			National Laboratory of Pattern Recognition (NLPR) \and
The Institute of Microelectronics, Tsinghua University (THU) \and
Chinese Academy of Sciences Artificial Intelligence Research (CAS-AIR) \\
\email hongyuan.yu@cripac.ia.ac.cn plh17@mails.tsinghua.edu.cn {\{yhuang,wangliang\}@nlpr.ia.ac.cn}}
\maketitle              
    \newcommand\blfootnote[1]{%
    \begingroup 
    \renewcommand\thefootnote{}\footnote{#1}%
    \addtocounter{footnote}{-1}%
    \endgroup 
    }
    
	
\begin{abstract}
This paper proposes a novel model for video generation and especially makes the attempt to deal with the problem of video generation from text descriptions, i.e., synthesizing realistic videos conditioned on given texts.
Existing video generation methods cannot be easily adapted to handle this task well, due to the frame discontinuity issue and their text-free generation schemes. To address these problems, we propose a recurrent deconvolutional generative adversarial network (RD-GAN), which includes a recurrent deconvolutional network (RDN) as the generator and a 3D convolutional neural network (3D-CNN) as the discriminator. The RDN is a deconvolutional version of conventional recurrent neural network, which can well model the long-range temporal dependency of generated video frames and make good use of conditional information.
The proposed model can be jointly trained by pushing the RDN to generate realistic videos so that the 3D-CNN cannot distinguish them from real ones. We apply the proposed RD-GAN to a series of tasks including conventional video generation, conditional video generation, video prediction and video classification, and demonstrate its effectiveness by achieving well performance.

\end{abstract}
%
%
%
\section{Introduction}

Image generation has drawn much attention recently, which focuses on synthesizing static images from random noises or semantic texts. But video generation, i.e., synthesizing dynamic videos including sequences of static images with temporal dependency inside, has not been extensively studied. In this work, we wish to push forward this topic by generating better videos and dealing with a rarely investigated task of text-driven video generation.

The task here can be defined as follows: given a text describing a scene in which someone is doing something, the goal is to generate a video with similar content. This is challenging since it involves language processing, visual-semantic association and video generation together. A straightforward solution is to generalize the existing video generation models for this task. But it is not optimal due to the following reasons: 1) conventional video generation directly generates videos from noises but not from semantic texts as our case, and 2) existing video generation models mostly suffer from the visual discontinuity problem, since they either directly ignore the modeling of temporal dependency of generated videos, or simply consider it in a limited range with 3D deconvolution.

In this paper, we propose a recurrent deconvolutional generative adversarial network (RD-GAN) for conditional video generation. The proposed RD-GAN first represents given semantic texts as latent vectors with skip-thoughts~\cite{kiros2015skip}, and exploits a generator named recurrent deconvolutional network (RDN) to generate videos in a frame-by-frame manner based on the latent vectors. The RDN can be regarded as a deconvolutional version of conventional recurrent neural network by replacing all the full connections with weight-sharing convolutional and deconvolutional ones. Accordingly, its hidden states now are 2D feature maps rather than 1D feature vectors, which efficiently facilitates the modeling of spatial structural patterns. After the generation, the generated videos are then fed into a discriminator which uses 3D convolutional neural network (3D-CNN) to distinguish from non-generated real videos. The generator and discriminator in RD-GAN can be jointly trained with the goal to generate realistic videos that can confuse the discriminator.
To demonstrate the effectiveness of our proposed RD-GAN, we perform various experiments in terms of conventional and conditional video generation, video prediction and classification.

Our contributions are summarized as follows. To the best of our knowledge, we make the attempt to study the problem of sentence-conditioned video generation, which is a rarely investigated but very important topic for the current research interest on generative models. We propose a novel model named recurrent deconvolutional generative adversarial networks to deal with the task, which is demonstrated to achieve good performance in a wide range of both generative and discriminative tasks.
\section{Related Work}

\textbf{Image Generation}.
With the fast development of deep neural networks, the generative models have made great progress recently. As we know, Tijmen~\cite{tieleman2014optimizing} proposed capsule networks to generate images and Dosovitskiy et al.~\cite{dosovitskiy2015learning}  generated  3D  chairs,  tables  and cars with deconvolutional  neural  networks. There are also some other works~\cite{reed2015deep,yang2015weakly} using supervised methods to generate images. Recently, the unsupervised methods, such as variational auto-encoder (VAE)~\cite{kingma2013auto} and generative adversarial network (GAN)~\cite{goodfellow2014generative} have attracted much attention. Gregor et al.~\cite{gregor2015draw} found that they can generate simple images by imitating human painting based on the recurrent variational autoencoder and the attention
mechanism. The autoregressive models proposed by Oord et al.~\cite{van2016pixel} modeled the conditional distribution of the pixel space and also achieved good experimental results. Compared with other methods, GAN has a relatively better performance on such tasks and various models~ \cite{wang2016generative,wu2017gp,bin2017high,tan2017improved} based on GAN generated appealing synthetic images. Beside that, conditional image generation has also been extensively studied. At first, simple variables, attributes or class labels~\cite{mirza2014conditional,van2016conditional} were used to generate specific images. Furthermore, researchers try to use unstructured text to do this work. For instance, Reed et al.~\cite{reed2016generating} used the text descriptions and object location constraints to generate images with conditional PixelCNN. The later works~ \cite{reed2016generative,han2017stackgan} built upon conditional GAN yielded 64$\times$64 or larger scale images of birds and flowers from text descriptions successfully.

\textbf{Video Generation}.
Before we go into our video generation model, it is essential to review the recent advance related to video prediction and generation. ~\cite{kalchbrenner2016video,zhou2016learning} inspired us in terms of video continuity. Compared with video prediction, there is no context information in the video generation task.
Vondrick et al.~\cite{vondrick2016generating} first came up with a "violence" generative model which can  directly yield fixed length videos from 3D deconvolution. However, the 3D deconvolution  causes more serious loss of information than 1D and 2D deconvolution. In order to fix this problem, Saito Masaki et al.~\cite{TGAN2017} proposed TGAN which tried to find all the latent vectors of continuous frames with the thought that videos are composed of images. And they used 1D and 2D deconvolution to generate video frame by frame. Unfortunately, their results do not have good continuity as they expected. MoCoGAN~\cite{tulyakov2017mocogan} decomposing motion and content for video generation, but their model lack understanding of semantics.
In this work, we propose recurrent deconvolutional generative adversarial network (RD-GAN) including a recurrent deconvolutional network (RDN) to handle the current problems in video generation and well exploit the conditional constraints.

\begin{figure}
	\centering
	\includegraphics[scale=0.62]{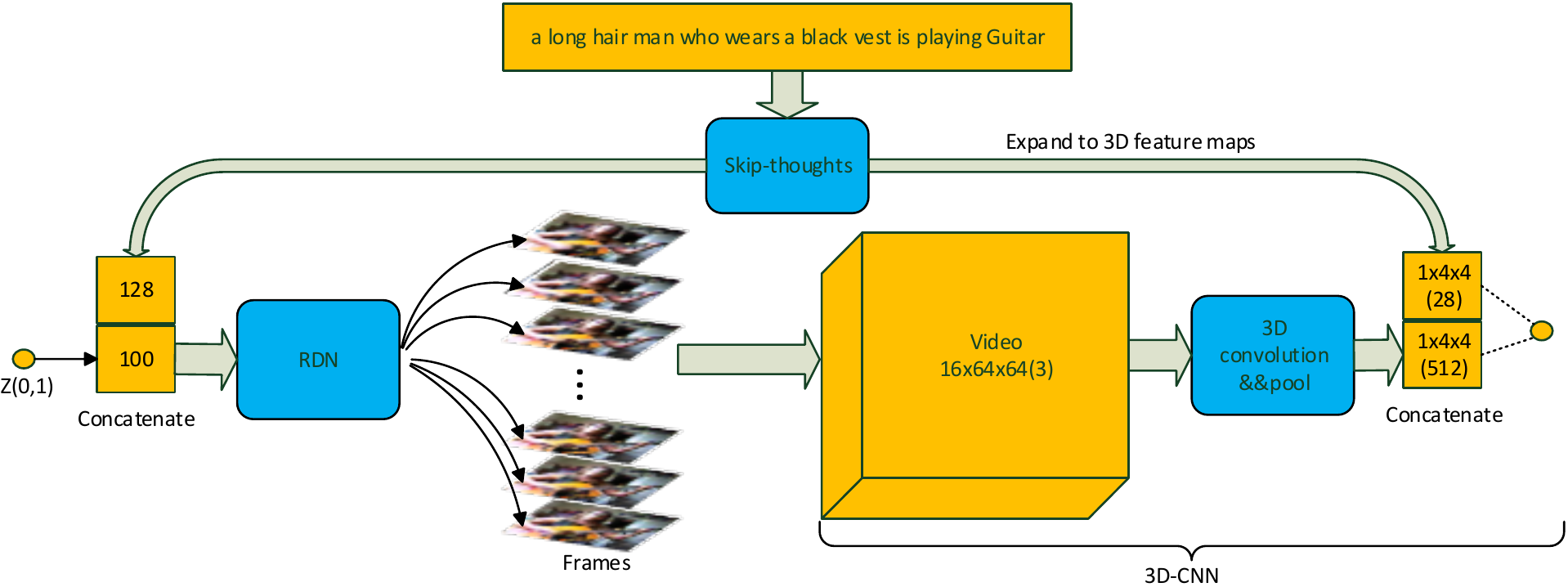}
	\caption{The proposed recurrent deconvolutional generative adversarial network (RD-GAN) for generating videos with text condition. The sentence is encoded into an 28-dimensional vector by skip-thoughts~\cite{kiros2015skip} and linear operation. In the generator, the text vector and an 100-dimensional vector sampled from Gaussian distribution are concatenated as input. Then the text vector is copied 4 times and the same 16 text vectors are stacked as 28$\times$1$\times$4$\times$4 feature maps before being fed into the penultimate layer of discriminator.}
	\label{fig:whole}
	\vspace{-0.5cm}
\end{figure}

\section{Recurrent Deconvolutional Generative Adversarial Network}
The architecture of our proposed recurrent deconvolutional generative adversarial network for generating videos with text condition is shown in Figure~\ref{fig:whole}. The RD-GAN is built upon the conventional generative adversarial network (GAN), which has two sub-networks: a generator and a discriminator. The generator is a recurrent deconvolutional network (RDN) which tries to generate more realistic videos, whose input is sequential concatenated vectors of noise sampled from Gaussian distribution and text embedding by skip-thoughts~\cite{kiros2015skip}. While the discriminator is a 3D convolutional neural network (3D-CNN) which tries to distinguish input videos between two classes: ``real'' and ``fake''. During the binary classification, it also exploits the text embedding that concatenated with video feature maps in the discriminator. These two sub-networks are jointly trained by playing a non-cooperative game, in which the generator RDN tries to fool the discriminator 3D-CNN while the discriminator aims to make few mistakes. Such a competitive scheme has been demonstrated to be effective for training generative models \cite{goodfellow2014generative}. In the following, we will present the generator and discriminator in details.

\subsection{Recurrent Deconvolutional Network as Generator}

The task of video generation based on semantic texts mainly encounters two challenges. One is how to extract suitable information from texts and associate it with the content of generated videos. It can be solved by exploiting recent advances in areas of natural language processing and multimodal learning. The other one is how to well model both long-range and short-range temporal dependencies of video frames, which focus on global slow-changing and local fast-changing patterns, respectively. The exist methods either just consider the local fast-changing ones with 3D convolutions or directly ignore the modeling of temporal dependency by treating each frame generation as a separate procedure.

\begin{figure*}
	\centering
	\includegraphics[scale=0.58]{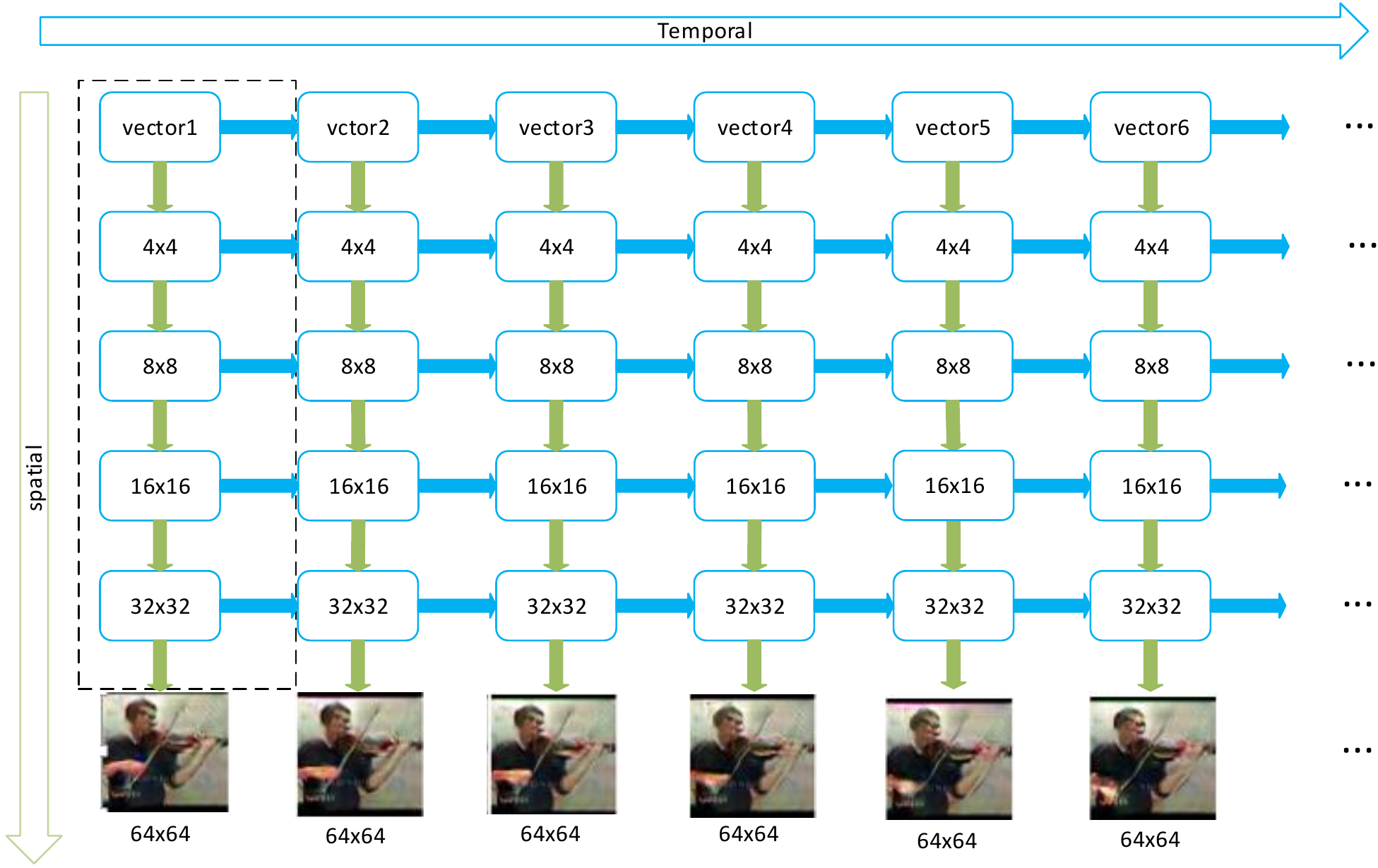}
	\caption{The proposed recurrent deconvolutional network as the generator. The input is vector1 and all vectors in the top level are 1D vectors connected by linear transformation. The rest are 2D feature maps connected by convolutions and we use blue arrows to represent them. While green arrows refer to deconvolution, which enlarges the scale of images in the spatial direction. The sizes of images and feature maps are annotated in the figure. All the different parameters are in the dashed box , and the rest are consistent with these parameters. (Best viewed in color)}
	\label{fig:rdn}
	\vspace{-0.5cm}
\end{figure*}

To well model the long-term temporal dependency during video generation, we propose a recurrent deconvolutional network (RDN) as shown in Figure~\ref{fig:rdn}. The RDN has a deep spatio-temporal architecture, whose input is a combination of noise and the vector extracted from the given text and the output is a sequence of generated video frames. Weight sharing is widely used in the temporal and the spatial direction, which can effectively reduce the number of parameters and contribute to model stability.

The whole network can be regarded as a nonlinear mapping from semantic texts to desired videos. In particular, in the spatial direction, there is a deep deconvolution network at each timestep for frame-wise generation. Between adjacent levels of feature maps, deconvolution, batch normalization and ReLU activation are successively used to upscale the frame size by a factor of 2.

In the temporal direction, it is a broad recurrent convolution network, in which pairwise feature maps at adjacent timesteps are connected by convolutions. It means that when predicting one video frame, the result can be directly modulated by its previous frame and recursively depends on other previous frames in a long temporal range. The formulation of inferring feature maps $H_{i,t}$ at the $i$-th level and the $t$-th timestep is:
\begin{equation}
\setlength{\abovedisplayskip}{5pt}
\setlength{\belowdisplayskip}{5pt}
H_{i,t} = a(b( H_{i-1,t} {\kern 2pt}  \hat* {\kern 2pt}  W_{i-1,t} + H_{i,t-1} * U_{i,t-1} + B_{i,t} ))
\end{equation}
where $*$ and $\hat*$ represent temporal convolution and spatial deconvolution operations, respectively. $U_{i,t-1}$ and $W_{i-1,t}$ contain the filter weights of temporal convolution and spatial deconvolution, respectively.
$B_{i,t}$ denotes the bias weights. $a(\cdot)$ and $b(\cdot)$ are ReLU activation function and batch normalization operation, respectively.
Due to the recurrent scheme, our RDN can flexibly produces videos with any length.

\subsection{Modified 3D Convolutional Neural Network as Discriminator}

The role of a discriminator is to identify the authenticity by classifying a given video into two classes: ``real'' and ``fake''. Considering that 3D convolutional neural network (3D-CNN) performs very well in the task of video classification, so we directly exploit it as our discriminator.

We make a few changes to the 3D-CNN as follows: 1) the last two linear layers are replaced by one 3D convolution layer, 2) the original sizes of some layers are reduced as the size of our video frame is 64$\times$64, 3) 3D batch normalization, leakyReLU activation function and 3D max-pooling are used after each 3D convolution, and 4) the text embedding by skip-thoughts is combined in the penultimate layer.


\subsection{Learning }



The weights of both generator and discriminator can be jointly trained by using the following objective:
\begin{equation}
\min \limits_{\theta_{G}}\max \limits_{\theta_{D}} \mathbb{E}_{x\sim p_{data}(x)}[\mathrm{log} D(x|t;\theta_{D})] + \\
  \mathbb{E}_{z\sim p_{z}(z)}[\mathrm{log} (1-D(G(z;\theta_{G})|t;\theta_{D}))]
\end{equation}
where $\theta_{G}$ and $\theta_{D}$ represent the parameters of generator and discriminator, respectively. $ p_{data}(x) $ is the distribution of real videos $x$. $z$ denotes the noise which is sampled from Gaussian distribution and $t$ is the text condition. The object will reach the global optimum when $ p_{G(z)} = p_{data} $. However, it is usually infeasible to obtain the global optimum, so we usually use gradient-based methods such as stochastic gradient descent (SGD)~\cite{nemirovski1978cezari} to find a good local equilibrium.

All the network weights are initialized by sampling from a Gaussian distribution with a mean of 0 and standard deviation of 0.02.  We use the ADAM solver to optimize all the parameters, in which the learning rate is 0.0002 and the momentum is 0.5. To speed up the training procedure, we first use all the images in video segments to train an image-based GAN, which has the same architecture in the spatial direction as the proposed RD-GAN. Then we use the learned weights as pretrained weights for the RD-GAN and fine-tune all the weights on videos. In fact, we can alternatively remove such a pretraining step with a longer training time.

\section{Experiments}

\subsection{Dataset and Implementation Details}

The UCF-101~\cite{soomro2012ucf101} dataset is used during the training process. It contains 13,320 videos belonging to 101 different classes of human actions. Because there are many classes of videos in the UCF-101, it is very difficult for our model to learn such a complex data distribution. As a result, we use videos belonging to the same class to train a separate network every time. Note that the whole UCF-101 dataset is used when we evaluate the representations taken from the discriminator.

To enlarge our training dataset, we divide each video into multiple video segments containing 16 consecutive frames. For example, frames 1 to 16 make up the first video segment, the second video segment consists of frames 2 to 17 and the third video segment consists of frames 3 to 18 ,and so on. So a single class of video can get about 20,000 video segments and the included frames are all resized to the size of 64$\times$64.

For the semantic texts associated with videos, we make the attempt to give different human-written text descriptions for different videos. For example, ``a cute little boy who wears a red headband and a black shirt is playing Violin'' or ``a curly young man who wears a black T-shirt is playing Guitar''.  Videos with similar content and background are put into the same class and named with the same text description because those videos in the UCF-101 are divided from the same long video.

\subsection{Video Generation from Sentence}

\begin{figure*}
	\centering
	\includegraphics[scale=0.45]{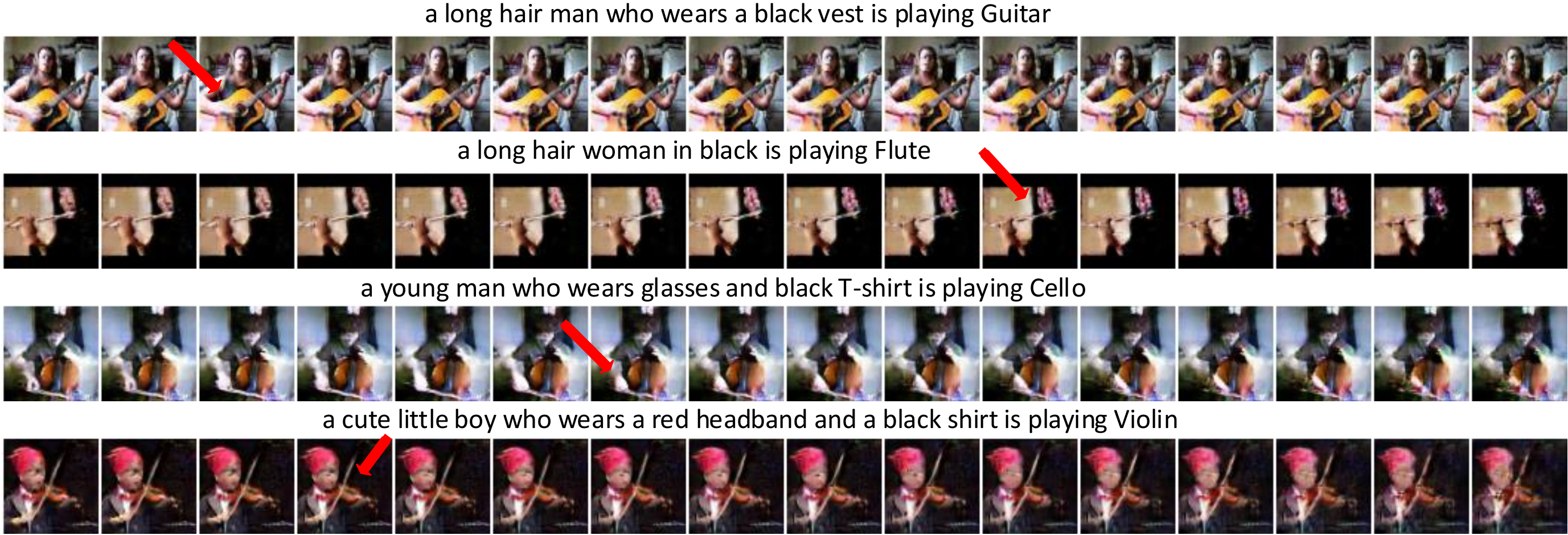}
	\caption{Results of conditional video generation. }
	\label{fig:im3}
	\vspace{-0.5cm}
\end{figure*}
\begin{figure*}
	\centering
	\includegraphics[scale=0.82]{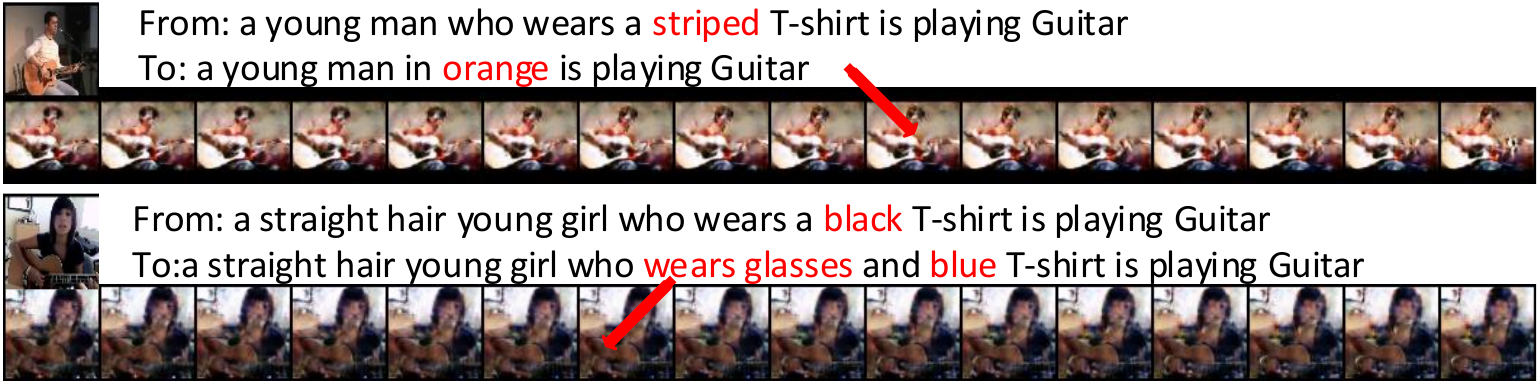}
	\caption{Results of conditional video generation with "unfamiliar" sentence . }
	\label{fig:trans}
	\vspace{-0.5cm}
\end{figure*}
Generating a specific video from human-written texts is equivalent to imagining a scene by our brain when we look at a novel. To adapt our model to the task of text-driven video generation, we exploit text information for both generator and discriminator as shown in Figure~\ref{fig:whole}. We use skip-thoughts \cite{kiros2015skip} to encode a given text into a 4800-dimensional vector, and then map the vector from 4800 to 28 dimensions. After that, we concatenate the 28-dimensional text vector with an 100-dimensional vector that is sampled from Gaussian distribution, then the new 128-dimensional vector is  used as the input of the generator. We also feed the same concatenated vector to the discriminator, which was extended to 3D data before combining with the penultimate layer of the 3D-CNN.

To simply annotate video data, we just gather similar videos into the same class and give them the same text descriptions.
Firstly, we generate videos from already known texts as shown in Figure~\ref{fig:im3}. From this figure we can see that our model is able to generate semantic-related videos. For example, when given the text: ``a long hair man who wears a black vest is playing Guitar'', our model can accordingly generate video with attributes: ``long hair'', ``man'' and ``black vest''. Then we try to change some attributes in sentence and send these "unfamiliar" sentences to our model. Figure~\ref{fig:trans} shows our model can generate new samples with the corect attributes. The result is not clear because the amount of data is not enough. In other words there is the problem of lack of continuous mapping between the semantic space and the image space.

\subsection{Video Classification}

\begin{table}
	\centering
	\caption{Accuracy of unsupervised methods on the UCF-101 dataset.}
	\begin{tabular}{lc}
		\midrule
		Method      \hspace{0cm}         & Accuracy   \\
		\midrule
		Chance                             & 0.9\%      \\
		STIP Features  ~\cite{soomro2012ucf101}                   & 43.9\%     \\
		Temporal Coherence  ~\cite{hadsell2006dimensionality}             & 45.4\%     \\
		Shuffle and Learn ~\cite{misra2016shuffle}               & 50.2\%     \\
		VGAN + Logistic Reg ~\cite{vondrick2016generating}              & 49.3\%     \\
		VGAN + Fine Tune  ~\cite{vondrick2016generating}              & 52.1\%     \\
		TGAN + Linear SVM    ~\cite{TGAN2017}       & 38.7\%     \\
		\textbf{Ours + Convolution Softmax}          & \textbf{53.3\%}     \\
		\textbf{Ours + Linear Softmax}               & \textbf{55.7\%}     \\
		\bottomrule
		ImageNet Supervision ~\cite{wang2015towards}             & 91.4\%
	\end{tabular}
	\label{tab:t1}
	\vspace{-0.5cm}
\end{table}

\begin{figure}[t]
	\centering
	\includegraphics[scale=0.45]{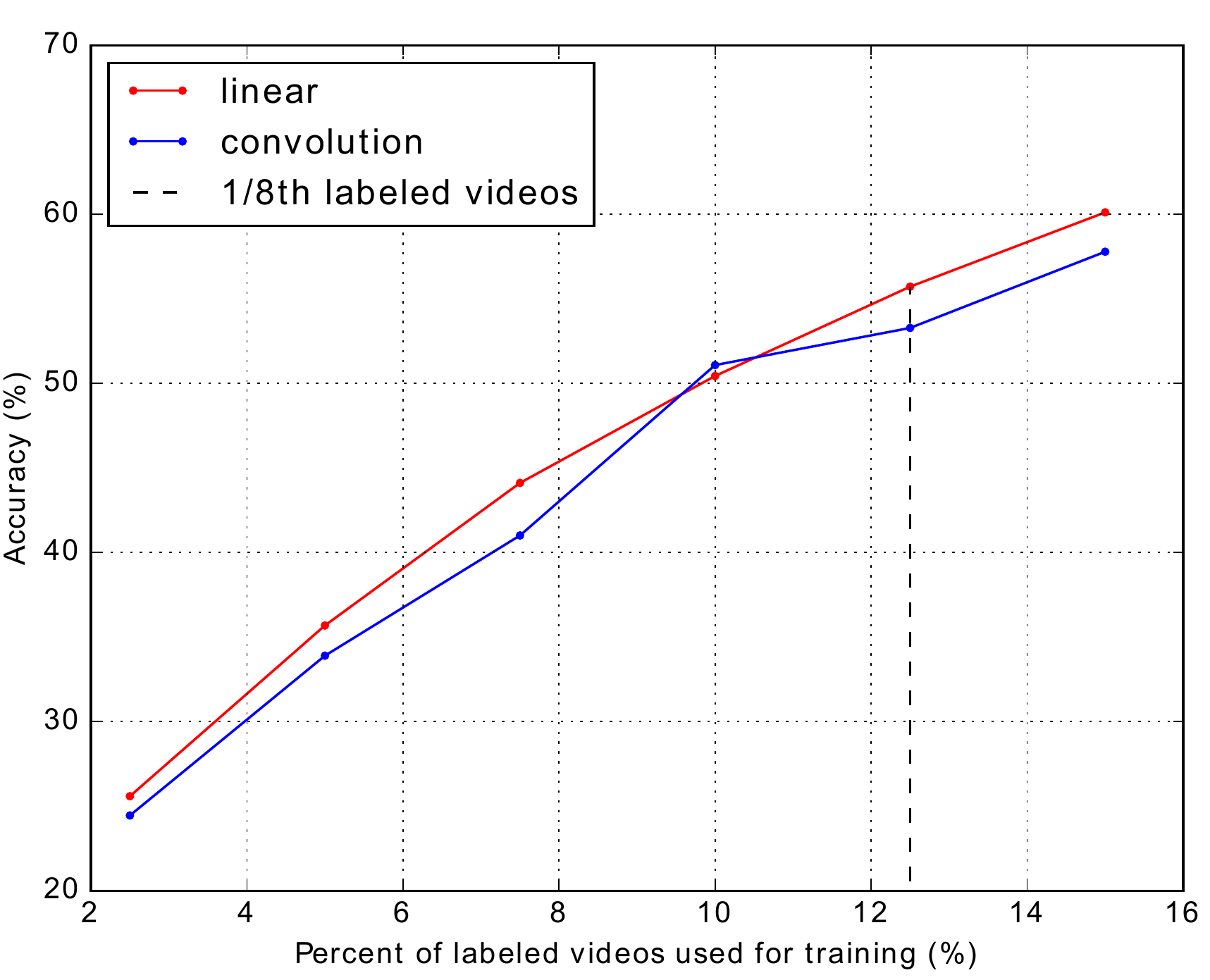}
	\caption{Ratio of labeled training videos vs Accuracy.}
	\label{tab:fen}
	\vspace{-0.5cm}
\end{figure}

Considering that the features in our discriminator are learned without supervision, we want to quantitatively analyze the capacity of discrimination on the task of video classification. Since the discriminator was originally used for binary classification, we have to replace its last layer with the softmax classifier for multi-class classification. All the videos in the UCF-101 dataset are used in this experiment. By segmenting all the videos into 16-frame segments, we can obtain totally 6 million video segments. The class for each video segment is in consistent with the class of its original video.

We first use the same softmax and dropout as \cite{vondrick2016generating} in the last two layers, which uses convolution to make the last feature maps transform into an 101-dimensional vector. We follow them and use 1/8 labeled videos to train the softmax classifier. Table~\ref{tab:t1} shows that our model improves the accuracy by 1.2\%. Considering that linear operation can also convert the last feature maps into an 101-dimensional vector for classification, we use liner operation to replace the previously used convolution. To our surprise, this liner operation further improves our performance by 2.4\%. Note that both convolution and linear operations have the same number of parameters. Obviously, models that leverage external supervision are still much better than unsupervised methods.
In Figure~\ref{tab:fen}, the performances of convolution softmax and linear softmax are also compared by using different numbers of labeled videos, and we can observe that the more number of training data, the higher the accuracy. On the whole, linear softmax produces more discirminative results than convolution softmax.

\section{Conclusions and Future Work}
We have proposed a recurrent deconvolutional generative adversarial network that generates videos based on semantic texts. Its generator is a deconvolutional version of recurrent neural network, which is able to well exploit the text information and efficiently models the long-range temporal dependency during video generation. We have performed various experiments in terms of conventional video generation, video generation driven by text, video prediction and video classification. The experimental results have demonstrated the effectiveness of our model.

Note that this is just an initial work on video generation, and it might have the following drawbacks. The proposed model based on GAN becomes unstable if trained with too many frames and currently cannot well generate clear videos of a larger size and a longer length. The processing of given text only involves simple steps of feature extraction. This might not be  optimal to associate it with diverse video contents. In the future, we will consider to train the RDN with more frames and extend our model to generate videos with a larger size and a longer length in a cascade manner.  Furthermore, we plan to expand the annotated videos in details for better text-driven video generation.

\section{ACKNOWLEDGMENTS}
This work is jointly supported by National Key Research
and Development Program of China (2016YFB1001000),
National Natural Science Foundation of China (61525306,
61633021, 61721004, 61420106015, 61806194), Capital Science and Technology
Leading Talent Training Project (Z181100006318030), 
Beijing Science and Technology Project (Z181100008918010) and CAS-AIR.

\bibliographystyle{splncs04}
\bibliography{mybibfile}

\end{document}